\documentclass[runningheads]{llncs}

\usepackage[T1]{fontenc}
\usepackage{amsmath}
\usepackage{amssymb}
\usepackage{booktabs}
\usepackage{graphicx}
\usepackage{multirow}
\usepackage{array}
\usepackage{url}
\usepackage{placeins}
\usepackage{float}
\usepackage{marvosym}
\usepackage[hidelinks]{hyperref}

\begin{document}

\title{DocDUR-Net: Degradation-Aware Unified Document Image Restoration}
\titlerunning{DocDUR-Net: Degradation-Aware Unified Document Image Restoration}

\author{Haowei Xu \and Peirong Zhang \and Xuhan Zheng \and Lianwen Jin(\Letter)}
\authorrunning{H. Xu et al.}

\institute{South China University of Technology, Guangzhou, China\\
\email{haowei\_xu@foxmail.com, \{eeprzhang,eexhzheng\}@mail.scut.edu.cn,\\
eelwjin@scut.edu.cn}}

\maketitle

\begin{abstract}
Captured document images often suffer from diverse degradations caused by uncontrolled imaging conditions, which reduce readability and obscure fine text structures. Single-degradation restoration methods are usually trained and deployed for individual degradation types, and therefore require model selection before restoration. Unified restoration models alleviate the need to maintain multiple restoration networks by sharing one network across tasks, but representative unified document image restoration methods still depend on external task selection. This dependency limits automatic document image restoration. We propose DocDUR-Net, a degradation-aware unified restoration network that derives restoration guidance directly from the input image. It introduces a Degradation Perception Module (DPM) to extract degradation representations and predict degradation weights, and Degradation Prior-guided Adaptive Feature Fusion (DP-AFF) to combine learnable degradation priors with degradation features extracted from the input. The resulting guidance is injected into a shared restoration backbone, enabling adaptive restoration without external task selection. DocDUR-Net also supports a manual mode with explicit task selection. Experiments on document image restoration benchmarks show that DocDUR-Net achieves the best average PSNR of 29.405 dB and SSIM of 0.960 among unified methods while remaining competitive with single-degradation methods.

\end{abstract}

\section{Introduction}

Document image restoration aims to recover readable document images from degraded captures~\cite{zhou2023review,zhang2025ocrgenbench}. In real-world acquisition scenarios, document images may suffer from shadows, blur, and specular highlights caused by uncontrolled illumination, camera motion, or reflective materials~\cite{ye2013document,wang2025wilddoc,zhang2026docaligner}. These degradations can obscure text regions and corrupt stroke structures, underscoring the need for effective restoration methods. 

\begin{figure}[t]
    \centering
    \includegraphics[width=1.0\textwidth]{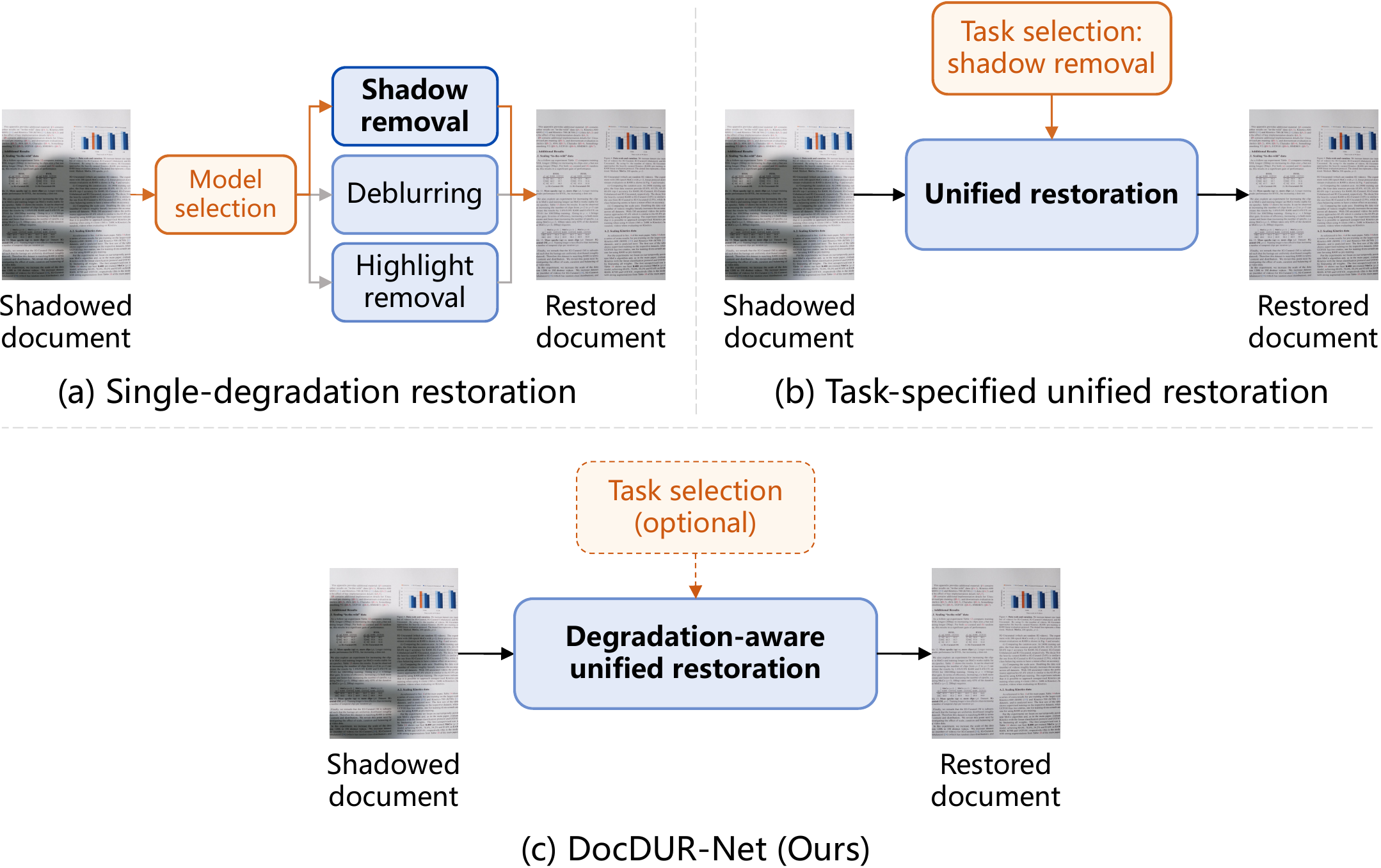}
    \caption{Comparison of document image restoration paradigms using shadow removal as an example. (a) Single-degradation restoration requires selecting a dedicated model for the target degradation. (b) Task-specified unified restoration uses one restoration model but still requires task selection. (c) DocDUR-Net extracts degradation information for automatic restoration while retaining explicit task selection in manual mode.}
    \label{fig:method_compare}
\end{figure}

Single-degradation document image restoration methods usually focus on one degradation type. Dedicated models are designed and trained for document shadow removal~\cite{lin2020bedsr,li2023high}, document image deblurring~\cite{hradivs2015convolutional,mei2019deepdeblur}, and document specular highlight removal~\cite{hou2021text,xu2025dochighlight}. While effective for their target degradations, this paradigm relies on separately trained models and requires model selection before restoration, as shown in Fig.~\ref{fig:method_compare}(a). Task-specified unified document image restoration methods share one model across tasks, reducing the need to train multiple restoration networks. However, existing unified methods still require explicit task selection before restoration, as shown in Fig.~\ref{fig:method_compare}(b). 

We therefore propose DocDUR-Net, a degradation-aware unified restoration network that performs automatic restoration by extracting degradation information from the input image, as shown in Fig.~\ref{fig:method_compare}(c). DocDUR-Net consists of a shared restoration backbone and two degradation-aware components. The Degradation Perception Module (DPM) extracts multi-scale degradation information and predicts degradation weights, while Degradation Prior-guided Adaptive Feature Fusion (DP-AFF) combines learnable degradation priors with the input degradation feature and injects the resulting guidance into the backbone. DocDUR-Net operates in automatic mode by default, while retaining a manual mode with explicit task selection for controlled evaluation and task-specified restoration.

Our contributions are summarized as follows:

\begin{itemize}
    \item We propose DocDUR-Net, a degradation-aware unified restoration method that enables automatic document image restoration by deriving restoration guidance from the input image, while retaining a manual mode with explicit task selection.
    \item We design DPM and DP-AFF to transform degradation perception into adaptive restoration guidance. DPM predicts degradation weights from the input image, and DP-AFF fuses learnable priors with input degradation features to guide the shared restoration backbone.
    \item Experiments on representative benchmarks demonstrate the effectiveness of DocDUR-Net. It achieves the best average PSNR and SSIM among unified methods while remaining competitive with single-degradation methods.
\end{itemize}

\section{Related Work}

\subsection{Single-Degradation Document Image Restoration}

Single-degradation document image restoration methods usually design specialized networks for a specific degradation. For document deblurring, representative methods~\cite{hradivs2015convolutional,mei2019deepdeblur} use large receptive fields or direct image regression to recover sharp character strokes. For document shadow removal, FSENet~\cite{li2023high} combines Transformer blocks~\cite{vaswani2017attention} with frequency decomposition, and DocDeshadower~\cite{zhou2024docdeshadower} removes shadows through attention aggregation and gated fusion. For document specular highlight removal, HAFNet~\cite{jiang2025hierarchical} adopts hierarchical adaptive filtering, while DocSHRNet~\cite{xu2025dochighlight} incorporates document structure and adaptive receptive fields. These methods are effective for their target degradations, but extending them to different restoration tasks requires separately trained models and model selection before inference.

Some document restoration frameworks use broadly applicable restoration architectures for different document restoration tasks~\cite{souibgui2022gan,yang2023docdiff,cicchetti2024naf,marulli2026docwavediff}. However, they are still trained and used separately for each task. This motivates unified document image restoration methods that handle different restoration tasks within a single model.

\subsection{Unified Document Image Restoration}

Unified modeling has been widely studied in document processing~\cite{tang2023udop,wan2024omniparser,zhang2025smaller}. In document image restoration, unified methods handle different restoration tasks with a shared model. DocRes~\cite{zhang2024docres} is a representative unified model that constructs dynamic visual prompts for different tasks and concatenates them with the degraded input to guide the restoration network. Uni-DocDiff~\cite{zhao2025unidocdiff} extends unified document restoration to diffusion models, using learnable task prompts to condition the iterative restoration process. These methods reduce the need to maintain separate restoration networks, but the restoration task is still selected externally for each input image. Their restoration process depends on externally specified task information rather than degradation information derived from the input image.

In contrast, DocDUR-Net incorporates degradation perception into unified restoration and guides restoration with degradation information extracted from the input image, while retaining a manual mode with explicit task selection.

\section{Method}

\subsection{Problem Formulation and Overview}

Given a degraded document image \(I_{\mathrm{input}}\), DocDUR-Net predicts a restored image \(\hat{I}\) using degradation-aware guidance \(g\):
\begin{equation}
    \hat{I}=\mathcal{R}(I_{\mathrm{input}}; g),
    \label{eq:degradation_guided_restoration}
\end{equation}
where \(\mathcal{R}(\cdot)\) denotes the restoration network and \(g\) is derived from degradation information in the input image.

\begin{figure}[t]
    \centering
    \includegraphics[width=1.0\textwidth]{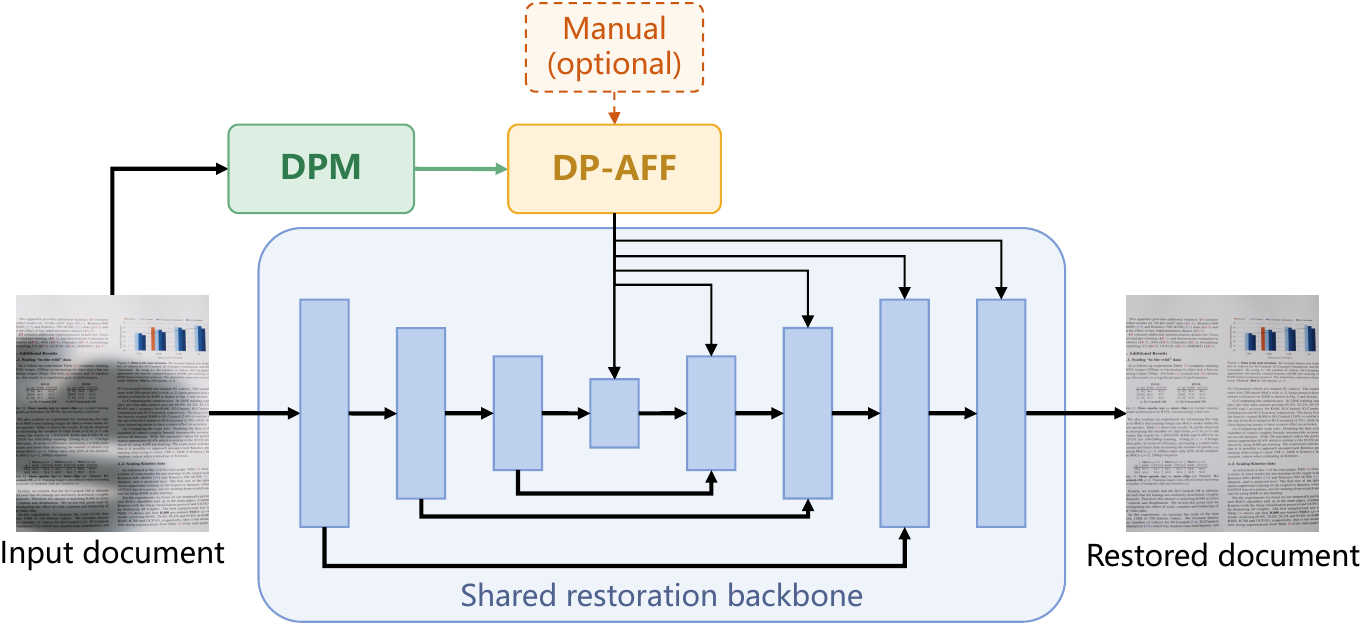}
    \caption{Overall architecture of DocDUR-Net. DPM predicts degradation weights from the input image, and DP-AFF injects degradation-guided features into the shared restoration backbone. The optional manual mode supports explicit task selection.}
    \label{fig:unified_arch}
\end{figure}

DocDUR-Net implements this formulation with a shared restoration backbone and two core components, as shown in Fig.~\ref{fig:unified_arch}. DPM extracts degradation information and predicts degradation weights in parallel with the backbone, while DP-AFF combines degradation information with learnable priors to produce feature guidance.

\subsection{Degradation Perception Module}

\begin{figure}[t]
    \centering
    \includegraphics[width=1.0\textwidth]{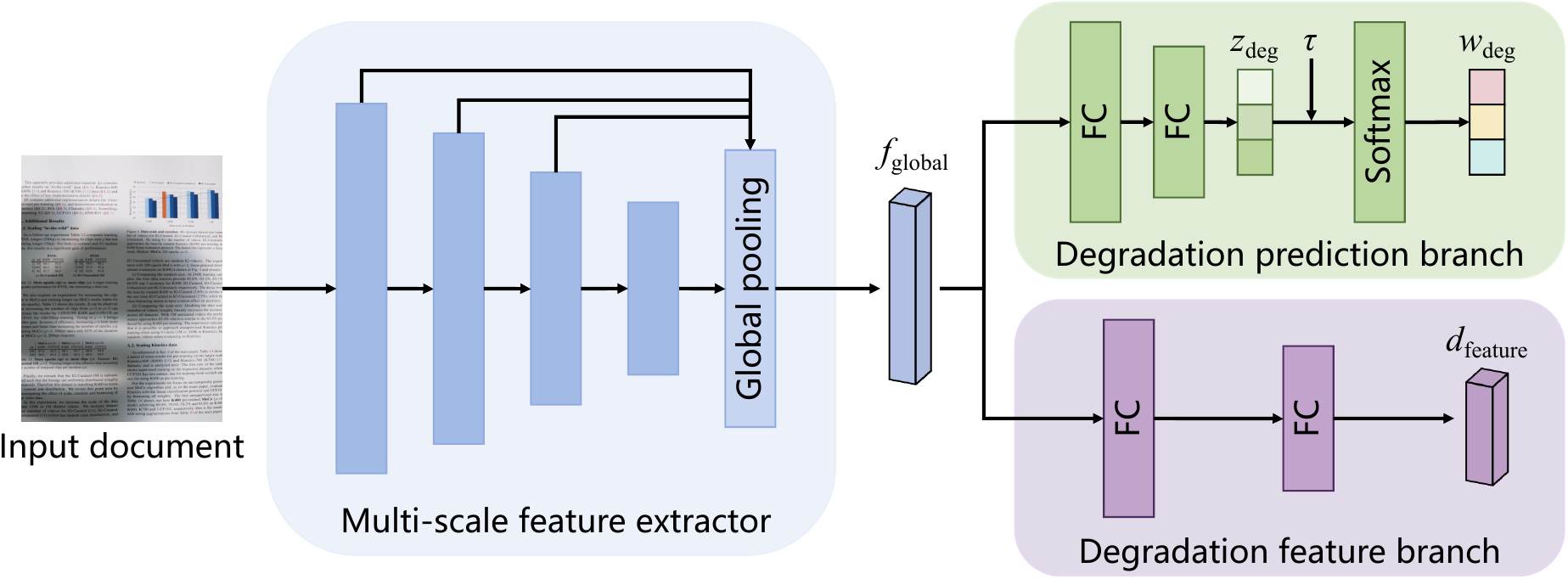}
    \caption{Structure of DPM for degradation feature extraction and weight prediction.}
    \label{fig:degradation_encoder}
\end{figure}

DPM extracts degradation information from the input image using a lightweight multi-stage encoder, as shown in Fig.~\ref{fig:degradation_encoder}. Features from four stages are compressed by global average pooling~\cite{lin2013network} and concatenated along the channel dimension to form a global degradation representation \(f_{\mathrm{global}}\).

DPM contains two parallel branches. The degradation prediction branch maps \(f_{\mathrm{global}}\) to degradation logits \(z_{\mathrm{deg}} \in \mathbb{R}^{K}\), where \(K\) denotes the number of restoration tasks. A learnable temperature parameter \(\tau\)~\cite{guo2017calibration} scales the logits before Softmax:
\begin{equation}
    w_{\mathrm{deg}} = \mathrm{Softmax}\left(\frac{z_{\mathrm{deg}}}{\tau}\right),
    \label{eq:task_weights}
\end{equation}
where \(w_{\mathrm{deg}} \in \mathbb{R}^{K}\) denotes the degradation weights used by DP-AFF to aggregate degradation priors. The degradation feature branch projects \(f_{\mathrm{global}}\) into a continuous degradation feature:
\begin{equation}
    d_{\mathrm{feature}} = f_{\mathrm{proj}}(f_{\mathrm{global}}),
    \label{eq:degradation_vector}
\end{equation}
where \(f_{\mathrm{proj}}(\cdot)\) is implemented by two fully connected layers and \(d_{\mathrm{feature}} \in \mathbb{R}^{d}\).

\subsection{Degradation Prior-guided Adaptive Feature Fusion}

\begin{figure}[t]
    \centering
    \includegraphics[width=0.9\textwidth]{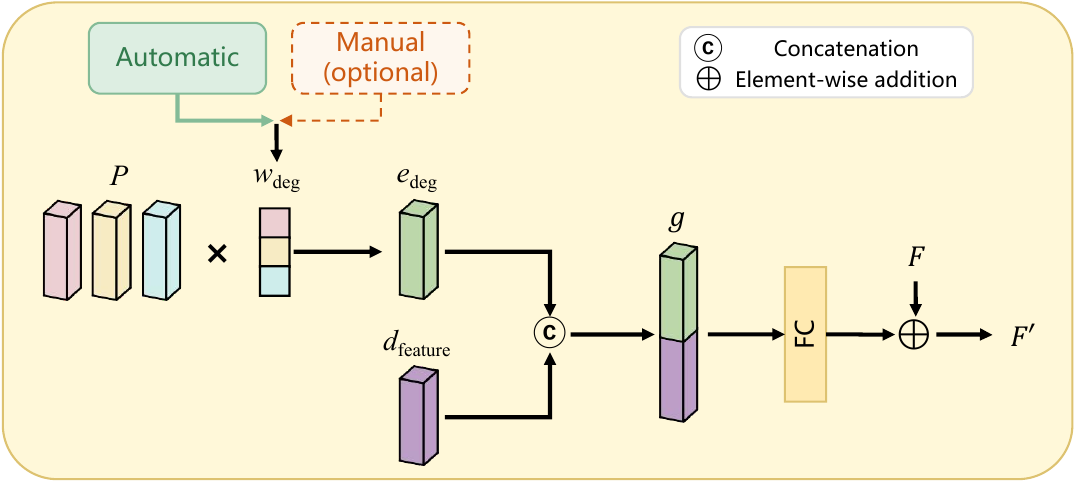}
    \caption{Structure of DP-AFF for degradation prior fusion and feature guidance.}
    \label{fig:dp_aff}
\end{figure}

DP-AFF uses the degradation weights to aggregate learnable priors and fuse them with the input degradation feature. The fused representation provides restoration guidance for the backbone, as shown in Fig.~\ref{fig:dp_aff}.

We define a learnable degradation prior matrix \(P \in \mathbb{R}^{d \times K}\), with each column corresponding to a restoration task. Given \(w_{\mathrm{deg}}\), DP-AFF obtains an aggregated degradation prior:
\begin{equation}
    e_{\mathrm{deg}} = P w_{\mathrm{deg}},
    \label{eq:task_prior_fusion}
\end{equation}
where \(e_{\mathrm{deg}} \in \mathbb{R}^{d}\). The prior \(e_{\mathrm{deg}}\) is concatenated with \(d_{\mathrm{feature}}\) and projected to form guidance:
\begin{equation}
    g = f_{\mathrm{fusion}}\left(\mathrm{Concat}[d_{\mathrm{feature}}, e_{\mathrm{deg}}]\right),
    \label{eq:guidance_generation}
\end{equation}
where \(f_{\mathrm{fusion}}(\cdot)\) is implemented by two fully connected layers. The vector \(e_{\mathrm{deg}}\) represents the weighted degradation prior, while \(d_{\mathrm{feature}}\) preserves input degradation information.

For a backbone feature \(F\), the projected guidance is added to \(F\):
\begin{equation}
    F' = F + \phi(g),
    \label{eq:feature_fusion}
\end{equation}
where \(\phi(\cdot)\) denotes the linear projection at the corresponding injection position, followed by spatial broadcasting. This additive fusion is applied across restoration stages to guide both global structure recovery and local detail refinement.

\subsection{Inference Modes}

DocDUR-Net primarily operates in automatic mode, as shown in Fig.~\ref{fig:unified_arch} and Fig.~\ref{fig:dp_aff}. In this mode, DPM predicts \(w_{\mathrm{deg}}\) from the input image, and DP-AFF uses it to aggregate the degradation prior and guide the backbone. This enables unified restoration without external task selection.

Manual mode serves as an optional protocol for controlled evaluation and restoration with explicit task selection. In this mode, the selected task is mapped to degradation weights in place of the DPM prediction, while the remaining restoration process is unchanged.

\subsection{Loss Function}

Training uses restoration and degradation classification supervision. The restoration loss is composed of pixel-level L1 loss, perceptual loss~\cite{johnson2016perceptual}, and SSIM loss~\cite{wang2004image}:
\begin{equation}
    \mathcal{L}_{\mathrm{rec}} =
    \lambda_{\mathrm{L1}}\mathcal{L}_{\mathrm{L1}} +
    \lambda_{\mathrm{perc}}\mathcal{L}_{\mathrm{perc}} +
    \lambda_{\mathrm{SSIM}}\mathcal{L}_{\mathrm{SSIM}}.
    \label{eq:loss_rec}
\end{equation}
The degradation prediction branch is supervised by the degradation label:
\begin{equation}
    \mathcal{L}_{\mathrm{cls}} = \mathrm{CE}(z_{\mathrm{deg}}, y_{\mathrm{deg}}),
    \label{eq:loss_cls}
\end{equation}
where \(y_{\mathrm{deg}}\) denotes the degradation label. The total objective is
\begin{equation}
    \mathcal{L}_{\mathrm{total}} =
    \mathcal{L}_{\mathrm{rec}} +
    \lambda_{\mathrm{cls}}\mathcal{L}_{\mathrm{cls}}.
    \label{eq:loss_total}
\end{equation}
We set \(\lambda_{\mathrm{L1}}=1\), \(\lambda_{\mathrm{perc}}=0.1\), \(\lambda_{\mathrm{SSIM}}=0.5\), and \(\lambda_{\mathrm{cls}}=0.1\).

\section{Experiments}

\subsection{Datasets and Tasks}

We use representative benchmarks for document shadow removal, document image deblurring, and document specular highlight removal. For document shadow removal, we use SD7K~\cite{li2023high} and follow the original split with 6479 training pairs and 760 testing pairs. For document image deblurring, we use TDD~\cite{hradivs2015convolutional}; following the split used by DocRes~\cite{zhang2024docres}, we train with 40000 image pairs and test with 1600 image pairs. For document specular highlight removal, we use DocHighlight~\cite{xu2025dochighlight}, with 1701 training pairs and 500 testing pairs from 2201 high-resolution real paired images. DocDUR-Net is trained as a single unified model on the union of all training sets.

\subsection{Evaluation Metrics}

We report PSNR and SSIM~\cite{wang2004image} between restored images and paired ground truth to measure pixel fidelity and structural similarity. For unified methods, Table~\ref{tab:main_results} also reports average scores over the evaluated tasks.

\subsection{Implementation Details}

All experiments are implemented with PyTorch~\cite{paszke2019pytorch}, using Restormer~\cite{zamir2022restormer} as the shared restoration backbone. Training is conducted on four NVIDIA RTX A6000 GPUs. We use AdamW~\cite{loshchilov2017decoupled} with weight decay \(1\times10^{-4}\). The degradation feature dimension and prior dimension are set to \(d=128\). The learnable temperature \(\tau\) is initialized to \(1.0\) and constrained to \([0.1,10.0]\).

To balance the different dataset sizes, each task dataset is sampled with equal probability during training. DocDUR-Net is trained in two stages. The first stage uses \(256\times256\) random crops for 100000 iterations with batch size 4 and learning rate \(2\times10^{-4}\). The second stage uses learning rate \(5\times10^{-5}\) for another 100000 iterations, with batch size 1, \(512\times512\) crops for DocHighlight and SD7K, and \(256\times256\) crops for TDD. The DPM input is resized to a 768-pixel long side, and random rotation is used for augmentation. For document specular highlight removal, we follow DocSHRNet~\cite{xu2025dochighlight} and use \(1024 \times 1024\) sliding-window inference with stride 512, while image-level inference is used for shadow removal and deblurring. 

\subsection{Comparison with Existing Methods}

Table~\ref{tab:main_results} compares DocDUR-Net with single-degradation and unified methods. The single-degradation methods include FSENet~\cite{li2023high} and DocDeshadower~\cite{zhou2024docdeshadower} for shadow removal, NAF-DPM~\cite{cicchetti2024naf} and DocWaveDiff~\cite{marulli2026docwavediff} for deblurring, and HAFNet~\cite{jiang2025hierarchical} and DocSHRNet~\cite{xu2025dochighlight} for document specular highlight removal. Unified methods include UPOCR~\cite{peng2024upocr} and DocRes~\cite{zhang2024docres}. We obtain the results of DocDeshadower, NAF-DPM, UPOCR, and DocRes by retraining the corresponding methods. The remaining competing results are taken from their original papers. UPOCR provides pixel-level outputs for document processing tasks, while DocRes uses task prompts for unified restoration. UPOCR and DocRes are evaluated with explicit task selection, whereas DocDUR-Net is evaluated in automatic mode.

\begin{table}[t]
    \centering
    \caption{Quantitative comparison on shadow removal, deblurring, and document specular highlight removal benchmarks. Best and second-best results in each task metric column are marked in bold and underlined. Average scores are reported for unified methods.}
    \label{tab:main_results}
    \setlength{\tabcolsep}{3pt}
    \renewcommand{\arraystretch}{1.2}
    \resizebox{\textwidth}{!}{%
    \begin{tabular}{@{}clcccccccc@{}}
        \toprule
        \multirow[c]{2}{*}{Setting} & \multirow[c]{2}{*}{Method} &
        \multicolumn{2}{c}{Shadow removal} & \multicolumn{2}{c}{Deblurring} &
        \multicolumn{2}{c}{Highlight removal} & \multicolumn{2}{c}{Average} \\
        \cmidrule(lr){3-4} \cmidrule(lr){5-6} \cmidrule(lr){7-8} \cmidrule(lr){9-10}
        & & PSNR & SSIM & PSNR & SSIM & PSNR & SSIM & PSNR & SSIM \\
        \midrule
        \multirow[c]{6}{*}{Single-degradation}
        & FSENet~\cite{li2023high} & 28.670 & 0.960 & -- & -- & -- & -- & -- & -- \\
        & DocDeshadower~\cite{zhou2024docdeshadower} & 25.024 & 0.917 & -- & -- & -- & -- & -- & -- \\
        & NAF-DPM~\cite{cicchetti2024naf} & -- & -- & \underline{26.940} & \underline{0.974} & -- & -- & -- & -- \\
        & DocWaveDiff~\cite{marulli2026docwavediff} & -- & -- & 26.280 & 0.973 & -- & -- & -- & -- \\
        & HAFNet~\cite{jiang2025hierarchical} & -- & -- & -- & -- & 28.848 & 0.928 & -- & -- \\
        & DocSHRNet~\cite{xu2025dochighlight} & -- & -- & -- & -- & \textbf{30.198} & \textbf{0.942} & -- & -- \\
        \midrule
        \multirow[c]{3}{*}{Unified}
        & UPOCR~\cite{peng2024upocr} & \underline{29.852} & \underline{0.962} & 23.850 & 0.951 & 28.576 & 0.914 & 27.426 & 0.942 \\
        & DocRes~\cite{zhang2024docres} & 29.071 & \textbf{0.965} & 25.554 & 0.966 & 29.385 & 0.929 & \underline{28.003} & \underline{0.953} \\
        & Ours & \textbf{30.318} & \textbf{0.965} & \textbf{28.114} & \textbf{0.981} & \underline{29.783} & \underline{0.933} & \textbf{29.405} & \textbf{0.960} \\
        \bottomrule
    \end{tabular}}
\end{table}

Without external task selection, DocDUR-Net achieves the highest average PSNR and SSIM among unified methods, with scores of 29.405 dB and 0.960. On shadow removal and deblurring, it also outperforms the single-degradation methods in both PSNR and SSIM. On specular highlight removal, DocSHRNet obtains the highest PSNR and SSIM, followed by DocDUR-Net, which surpasses HAFNet and the other unified methods. These results demonstrate the effectiveness of DocDUR-Net for unified document image restoration.

\begin{figure}[!htbp]
    \centering
    \includegraphics[width=1.0\textwidth]{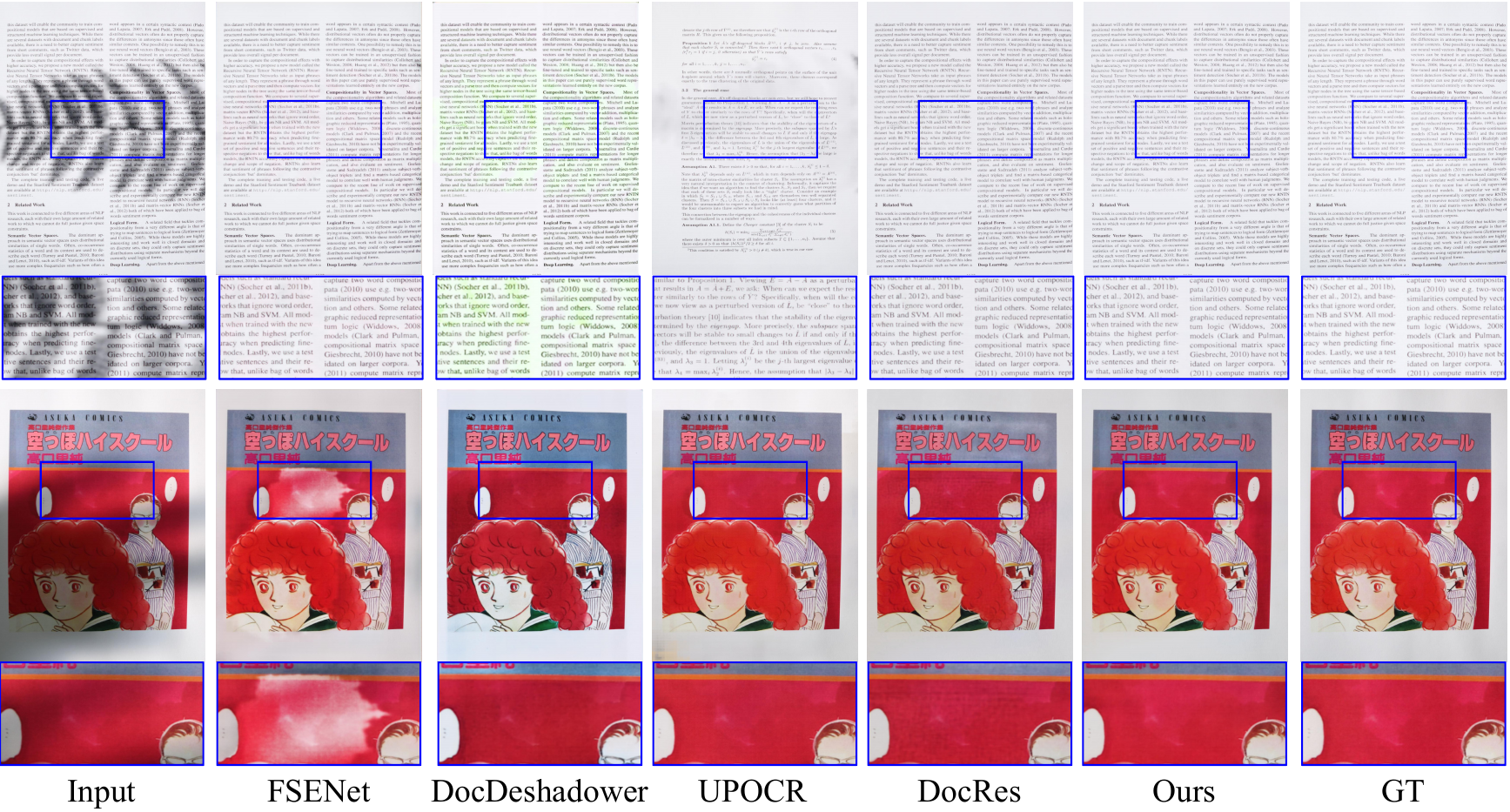}
    \caption{Qualitative comparison on document shadow removal.}
    \label{fig:visual_shadow}
\end{figure}

\begin{figure}[!t]
    \centering
    \includegraphics[width=1.0\textwidth]{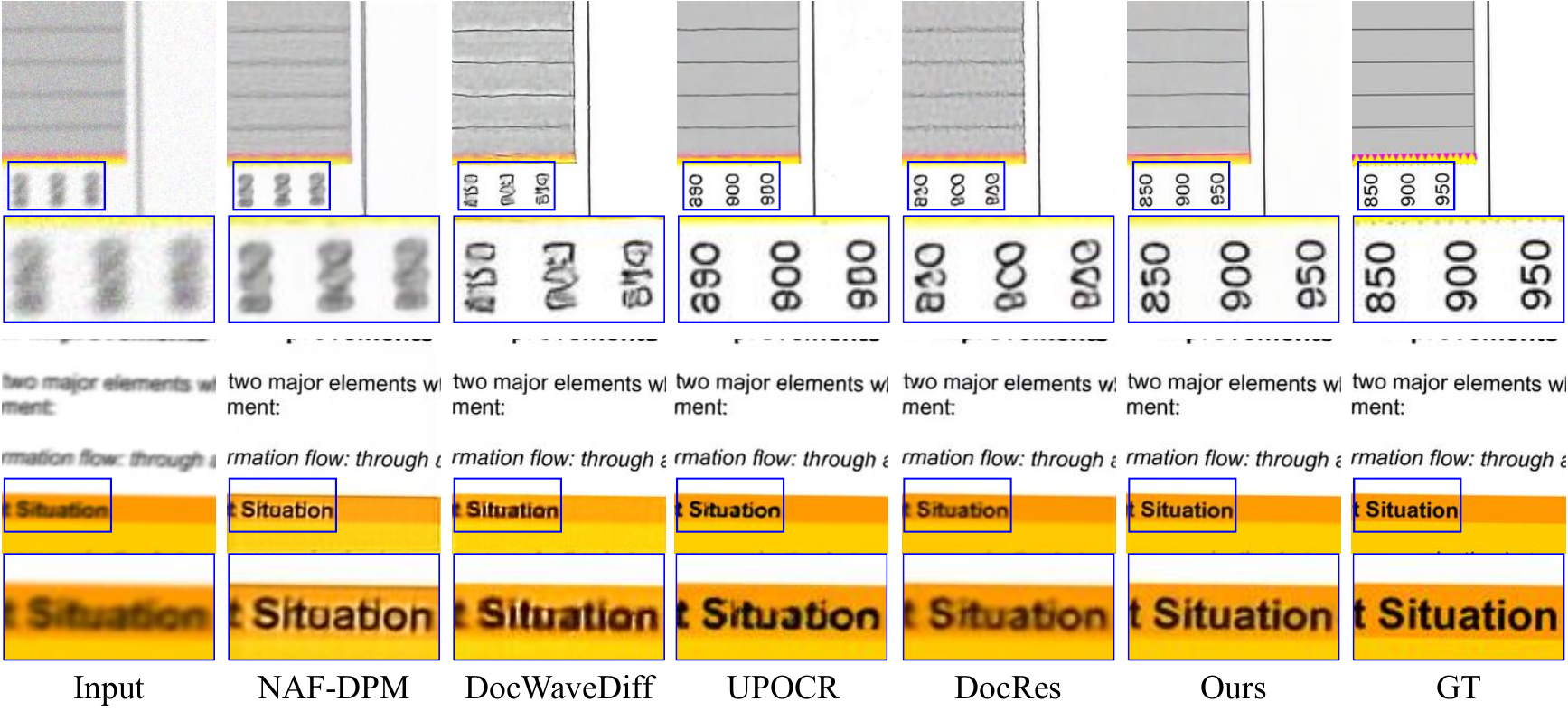}
    \caption{Qualitative comparison on document image deblurring.}
    \label{fig:visual_blur}
\end{figure}

\begin{figure}[!t]
    \centering
    \includegraphics[width=1.0\textwidth]{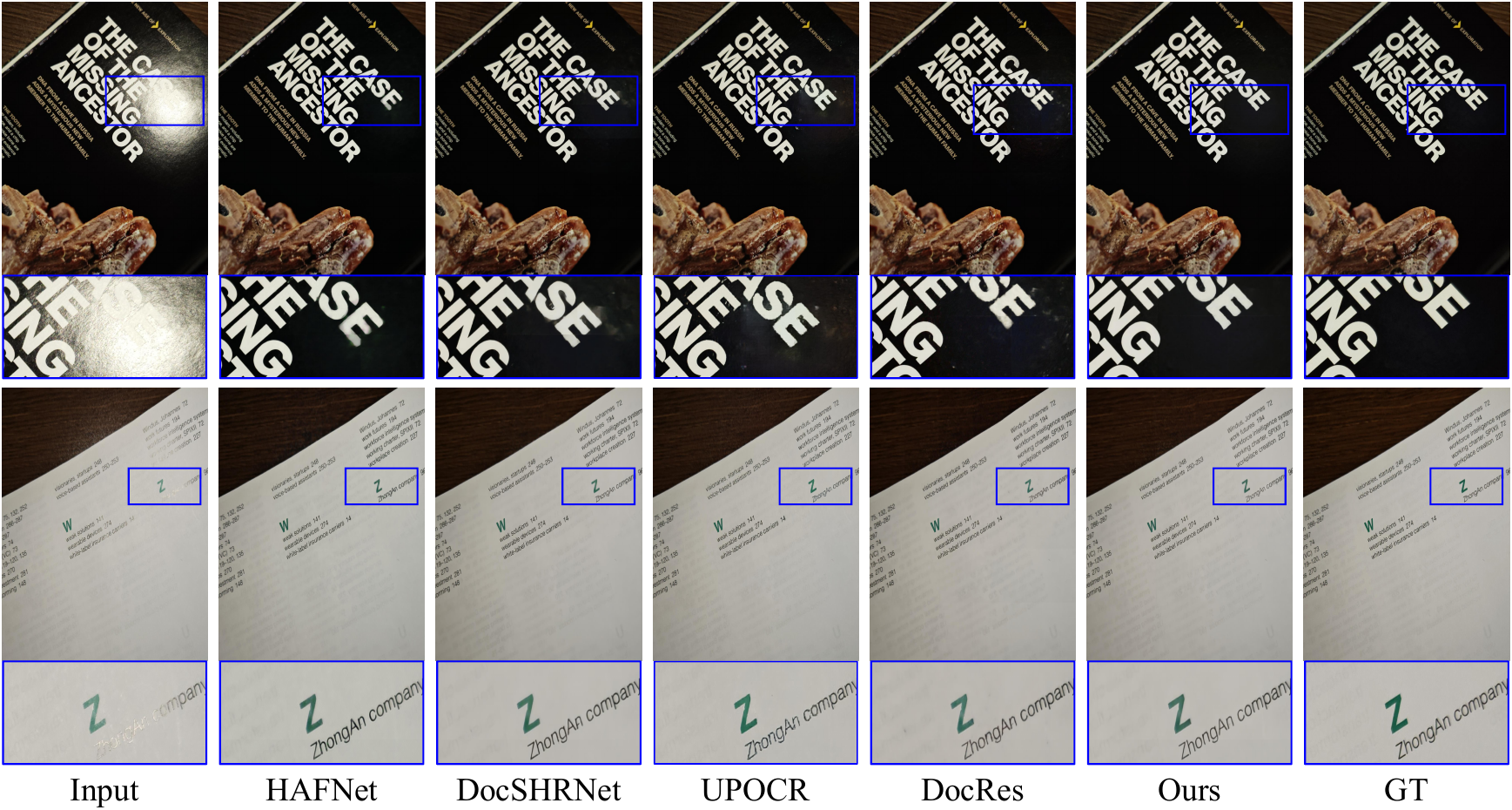}
    \caption{Qualitative comparison on document specular highlight removal.}
    \label{fig:visual_highlight}
\end{figure}

Figs.~\ref{fig:visual_shadow}, \ref{fig:visual_blur}, and \ref{fig:visual_highlight} show qualitative comparisons on the three restoration tasks. DocDUR-Net removes shadows while preserving document details, reconstructs sharper character strokes for deblurring, and suppresses specular highlights with fewer visible artifacts.

% \FloatBarrier

\subsection{Ablation Study}

Table~\ref{tab:ablation} studies the contribution of DPM and DP-AFF. The baseline is the shared restoration backbone trained jointly on all tasks without explicit degradation perception. With DPM alone, the degradation feature is directly projected and added to backbone features. DPM improves the average PSNR from 28.144 dB to 29.256 dB, indicating that input degradation information is useful for unified restoration. DP-AFF further improves the average PSNR to 29.405 dB by combining input degradation features with learnable degradation priors. The full model improves the average PSNR by 1.261 dB with only a 2.3\% parameter increase.
\begin{table}[t]
    \centering
    \caption{Ablation results for DPM and DP-AFF.}
    \label{tab:ablation}
    \renewcommand{\arraystretch}{1.3}
    \setlength{\tabcolsep}{3pt}
    \resizebox{\textwidth}{!}{%
    \begin{tabular}{@{}ccccccccccc@{}}
        \toprule
        \multicolumn{2}{c}{Module} & \multicolumn{2}{c}{Shadow removal} & \multicolumn{2}{c}{Deblurring} &
        \multicolumn{2}{c}{Highlight removal} & \multicolumn{2}{c}{Average} & \multirow{2}{*}{Params (M)} \\
        \cmidrule(lr){1-2} \cmidrule(lr){3-4} \cmidrule(lr){5-6} \cmidrule(lr){7-8} \cmidrule(lr){9-10}
        DPM & DP-AFF & PSNR & SSIM & PSNR & SSIM & PSNR & SSIM & PSNR & SSIM & \\
        \midrule
        -- & -- & 28.931 & 0.964 & 27.699 & 0.980 & 27.804 & 0.923 & 28.144 & 0.956 & 26.13 \\
        \(\checkmark\) & -- & 30.193 & 0.965 & 27.871 & 0.980 & 29.706 & 0.933 & 29.256 & 0.959 & 26.69 \\
        \(\checkmark\) & \(\checkmark\) & \textbf{30.318} & \textbf{0.965} & \textbf{28.114} & \textbf{0.981} & \textbf{29.783} & \textbf{0.933} & \textbf{29.405} & \textbf{0.960} & 26.74 \\
        \bottomrule
    \end{tabular}}
\end{table}

\subsection{Analysis of Automatic and Manual Modes}

Table~\ref{tab:classification_acc} evaluates the degradation weights predicted by DPM. DPM matches all shadow removal and deblurring labels and achieves 99.4\% accuracy on document specular highlight removal, reaching 99.9\% overall accuracy. This result supports the reliability of automatic mode.
\begin{table}[!htbp]
    \centering
    \caption{Degradation perception accuracy of DPM.}
    \label{tab:classification_acc}
    \renewcommand{\arraystretch}{1}
    \setlength{\tabcolsep}{4pt}
    \begin{tabular}{@{}ccccc@{}}
        \toprule
        Task & Samples & Correct & Misclassified & Accuracy (\%) \\
        \midrule
        Shadow removal & 760 & 760 & 0 & 100.0 \\
        Deblurring & 1600 & 1600 & 0 & 100.0 \\
        Specular highlight removal & 500 & 497 & 3 & 99.4 \\
        \midrule
        Total & 2860 & 2857 & 3 & 99.9 \\
        \bottomrule
    \end{tabular}
\end{table}

We further compare automatic and manual modes. In automatic mode, the predicted degradation weights are used to construct the restoration guidance, whereas manual mode uses weights determined by the selected task. The results are identical on shadow removal and deblurring, with only a marginal PSNR difference of 0.071 dB on document specular highlight removal. This difference is consistent with Table~\ref{tab:classification_acc}, where all misclassified samples come from the specular highlight removal test set.

\begin{figure}[t]
    \centering
    \includegraphics[width=0.7\textwidth]{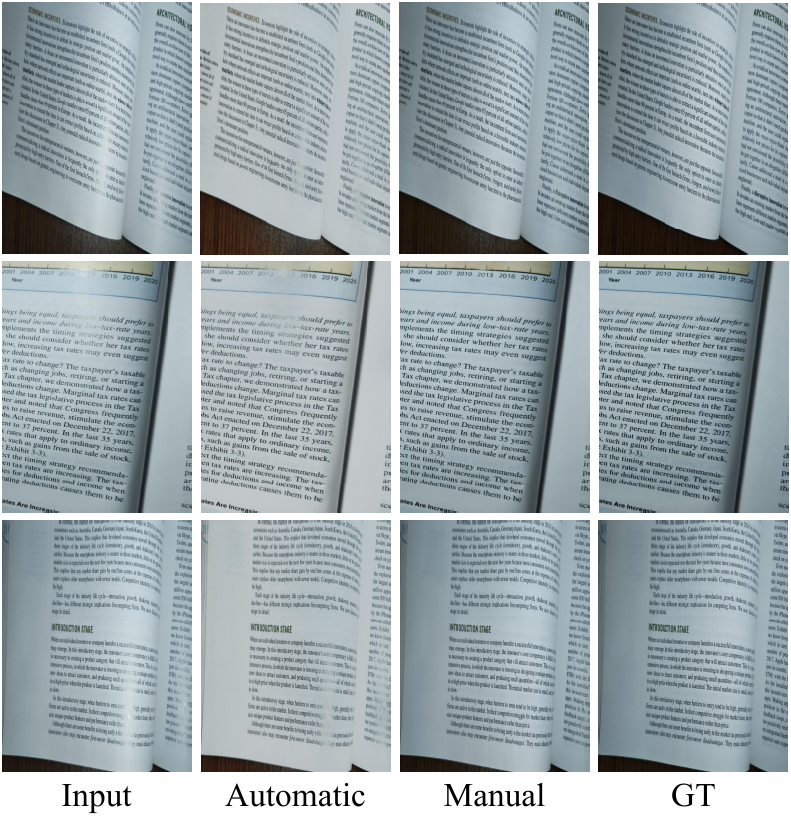}
    \caption{Visual comparison of automatic and manual modes for the three specular highlight removal test samples assigned to shadow removal by DPM.}
    \label{fig:misclassified_comparison}
\end{figure}

Fig.~\ref{fig:misclassified_comparison} visualizes all three specular highlight removal test samples assigned to shadow removal by DPM. These samples also contain visible shadow regions. In automatic mode, DocDUR-Net suppresses the shadowed areas while preserving readable text. With explicit selection of specular highlight removal, manual mode more directly suppresses reflective regions. This comparison shows that automatic mode remains reliable without external task selection, while manual mode provides flexible restoration control.

\section{Conclusion}

This paper presents DocDUR-Net, a degradation-aware unified document image restoration method. It extracts degradation information from the input image and uses DPM and DP-AFF to guide a shared restoration backbone, enabling automatic restoration without external task selection. The method also retains a manual mode for flexible restoration control. Experimental results demonstrate the effectiveness of DocDUR-Net for degradation-aware unified document image restoration.

\begin{credits}
\subsubsection{\ackname}
This research is supported in part by the National Natural Science Foundation of China (Grant No. 62476093), the Natural Science Foundation of Guangdong Province (Grant No. 2026A1515012038), the China Postdoctoral Science Foundation (Grant No. 2026M791625), and the Postdoctoral Fellowship Program (Grade B) of the China Postdoctoral Science Foundation (Grant No. GZB20260386).

\subsubsection{\discintname}
The authors have no competing interests to declare that are relevant to the content of this article.
\end{credits}

\bibliographystyle{splncs04}
\bibliography{references}

\end{document}